\documentclass[conference]{IEEEtran}

\usepackage{bibentry}  
\usepackage{xcolor,soul,framed} 

\colorlet{shadecolor}{yellow}
\usepackage[pdftex]{graphicx}
\graphicspath{{../pdf/}{../jpeg/}}
\DeclareGraphicsExtensions{.pdf,.jpeg,.png}

\usepackage[cmex10]{amsmath}
\usepackage{array}
\usepackage{mdwmath}
\usepackage{mdwtab}
\usepackage{eqparbox}
\usepackage{url}
\usepackage{graphicx} 
\usepackage{caption} 
\usepackage{hyperref}       
\usepackage{url}            
\usepackage{booktabs}       
\usepackage{amsfonts}       
\usepackage{nicefrac}       
\usepackage{microtype}      
\usepackage{xcolor}         
\usepackage{graphicx}
\usepackage{subfigure}
\usepackage{adjustbox}
\usepackage{makecell}
\usepackage{amsmath}
\usepackage{algorithm}
\usepackage{bbm}
\usepackage[noend]{algpseudocode}
\usepackage{lipsum} 
\usepackage{tabularx}
\makeatletter
\def\algbackskip{\hskip-\ALG@thistlm}
\makeatother
\algrenewcommand\algorithmicrequire{\textbf{Input:}}
\algrenewcommand\algorithmicensure{\textbf{Output:}}
\usepackage{booktabs}
\usepackage{hyperref}
\DeclareMathOperator*{\argmin}{argmin}
\def\BibTeX{{\rm B\kern-.05em{\sc i\kern-.025em b}\kern-.08em
    T\kern-.1667em\lower.7ex\hbox{E}\kern-.125emX}}

\usepackage{xspace}

\newcommand{\Name }{MetaEMS\xspace }

\begin{document}

    \title{Meta-Reinforcement Learning for Building Energy
Management System }
\author{
  Huiliang Zhang$^{1}$, Di Wu$^{1}$, Arnaud Zinflou$^{2}$, and Benoit Boulet$^{1}$\\
  $^{1}$ McGill University, Canada,
  \{huiliang.zhang2, di.wu5, benoit.boulet\}@mail.mcgill.ca\\
  $^{2}$Hydro-Québec Research Institute, Canada,
  zinflou.arnaud@hydroquebec.com
}

\maketitle

\begin{abstract}

The building sector is one of the largest contributors to global energy consumption. Improving its energy efficiency is essential for reducing operational costs and greenhouse gas emissions. Energy management systems (EMS) play a key role in monitoring and controlling building appliances efficiently and reliably. With the increasing integration of renewable energy, intelligent EMS solutions have received growing attention. Reinforcement learning (RL) has recently been explored for this purpose and shows strong potential. However, most RL-based EMS methods require a large number of training steps to learn effective control policies, especially when adapting to unseen buildings, which limits their practical deployment. This paper introduces \Name, a meta-reinforcement learning framework for EMS. \Name improves learning efficiency by transferring knowledge from previously solved tasks to new ones through group-level and building-level adaptation, enabling fast adaptation and effective control across diverse building environments. Experimental results demonstrate that \Name adapts more rapidly to unseen buildings and consistently outperforms baseline methods across various scenarios.

\end{abstract}
\begin{IEEEkeywords}
Building energy
management system, reinforcement learning,  meta-learning, smart grid.
\end{IEEEkeywords}
\section{Introduction}

Buildings currently account for 39\% of global greenhouse gas emissions and energy use, making them a critical focus for achieving both national and international climate goals. Moreover, as people spend over 85\% of their time indoors \cite{zhang2022building,raza2024smart}, effective building control strategies not only reduce emissions and energy costs, but also improve indoor comfort and occupant well-being. In response, building energy management systems (EMS) have received growing attention in recent years. EMS aims to optimize energy usage in smart buildings equipped with intelligent controls and automation. They can take advantage of dynamic electricity pricing and the increasing integration of distributed renewable energy resources by intelligently scheduling Energy Storage Units (ESU) and controllable loads such as Heating, Ventilation, and Air Conditioning (HVAC) systems \cite{raza2024smart,mariano2021review}.

%
Traditional control approaches, such as rule-based control (RBC) and model predictive control (MPC), require expert knowledge and are often limited to small-scale, static EMS settings \cite{mariano2021review,serale2018model}. These methods struggle to generalize across buildings with diverse structures and dynamic behaviors, leading to costly system redesigns in new deployments. Reinforcement learning (RL) offers a model-free alternative by learning control policies directly from interactions with the environment, enabling greater adaptability to uncertainties such as variable renewable generation and load demands \cite{arroyo2022reinforced,Multi-agent-havc,10677393}.

However, RL suffers from poor data efficiency, often requiring millions of interaction steps to achieve competitive performance \cite{10677393,NEURIPS2022_3908cadf,zhangreview,zeng2025acecoderacingcoderrl}. For instance, an RL agent may need up to 5 million interaction steps to match the performance of a traditional feedback controller for HVAC control \cite{forootani2022advanced}. Moreover, most RL-based methods assume static EMS environments, which is unrealistic given the diverse structures, usage patterns, and dynamics of real-world buildings \cite{10677393,chen2019gnu,KnnMTS,LMHR}. This assumption limits generalizability and often leads to suboptimal performance when deployed in unseen scenarios.
In addition to generalization challenges, EMS environments present further complexities. First, the underlying distributions of key variables—such as renewable generation, non-shiftable loads, outdoor temperature, and electricity prices—are often unknown. Second, ESU and HVAC systems are subject to temporally coupled operational constraints that vary across environments, making it difficult to train data-efficient control policies.

To address these challenges, we propose \Name, a meta-RL framework for EMS in buildings. \Name improves learning efficiency by transferring knowledge from previously solved tasks to learning tasks in unseen buildings. We formulate a learning-based energy optimization problem that manages ESU, HVAC systems, renewable energy, and non-shiftable loads, without requiring a known building dynamics model. Moreover, \Name minimizes energy costs while shaping load profiles to support grid reliability. To enhance adaptability and data efficiency, we introduce two adaptation mechanisms in \Name: group-level adaptation, which refines a shared initialization across tasks, and building-level adaptation, which fine-tunes parameters for each specific building control task. During training, each task inherits the shared initialization and performs localized adaptation, then contributes back to the group-level model.

The main contributions of this paper are: (i) We introduce \Name, a meta-RL framework for building energy management, 
which requires no prior knowledge of building dynamics and adapts quickly to new EMS scenarios with limited data.
To the best of our knowledge, this is the first work to apply meta-RL to EMS building control, addressing the key challenges of generalization, data efficiency, and dynamic operational constraints in real-world smart building environments.
(ii) We validate the effectiveness and practicality of \Name empirically across diverse real-world building scenarios. 
Experiment results show that \Name achieves faster convergence, stronger robustness on unseen buildings' control task, and better generalization across a variety of dynamic building environments.
\section{Related Work}
\subsection{Traditional Control Methods for EMS}
The traditional ways of building control can be sorted into RBC and MPC methods \cite{zhang2022building}.
The control strategies in RBC are manually designed based on expert knowledge and predefined thresholds. While simple to implement, RBC lacks adaptability and performs poorly in dynamic or large-scale scenarios. MPC improves upon RBC by leveraging system models to predict future states and optimize control actions over a time horizon \cite{mariano2021review,serale2018model}. MPC can effectively handle constraints and multi-variable systems, but its performance is highly dependent on the accuracy of the underlying building model, which is often hard to obtain in real-world settings. To address modeling challenges, data-driven MPC has emerged, where system dynamics are learned from historical data using machine learning techniques \cite{arroyo2022reinforced,10315171,10531060,dmg}. However, data-driven MPC still requires reliable data for model training and remains sensitive to distribution shifts across different building environments, limiting its generalizability and robustness.

\begin{figure*}[t]
\centering
 \setlength{\belowcaptionskip}{-0.3cm}
\includegraphics[width=0.6\linewidth]{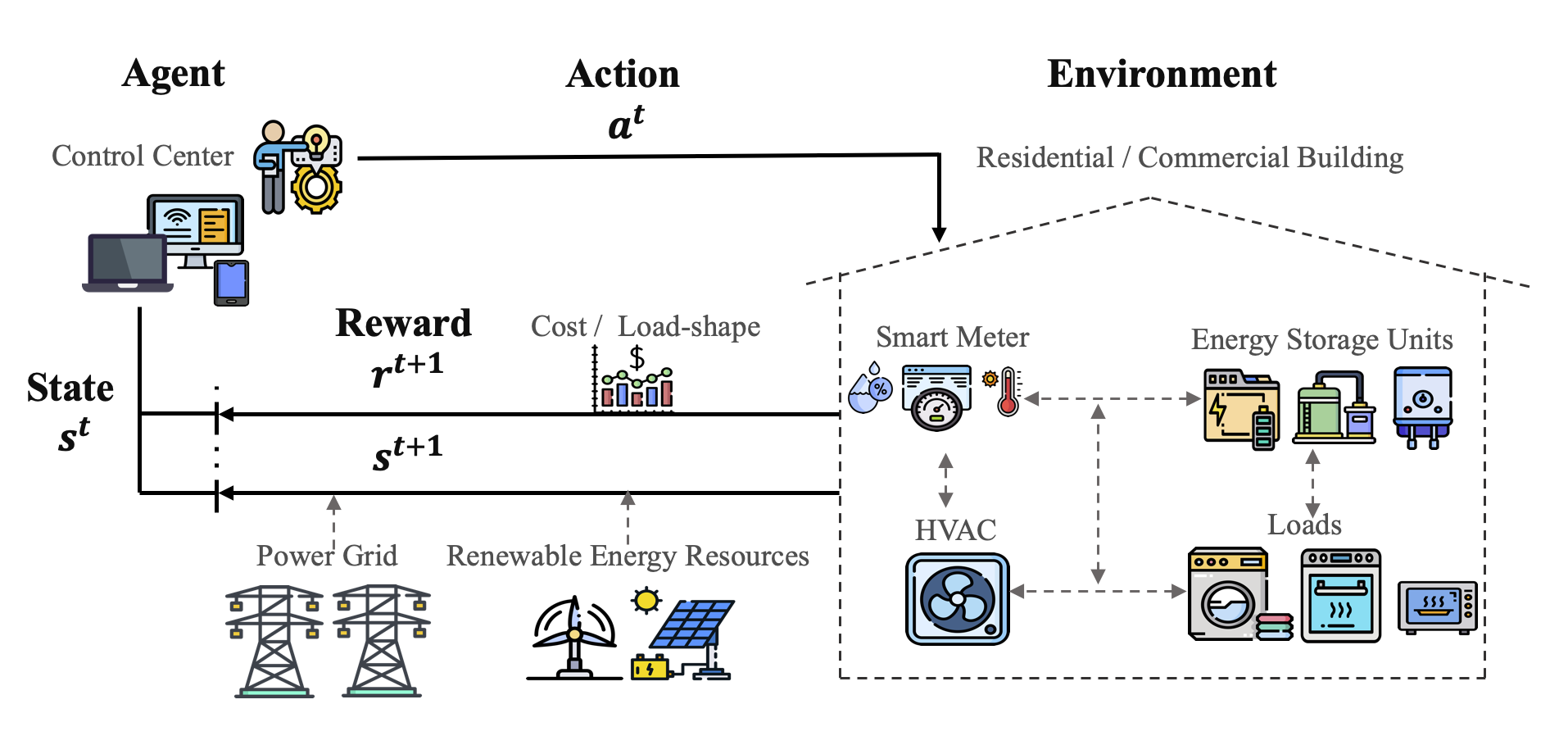}
\caption{A general diagram for RL with key components in EMS. A  typical  EMS  architecture  consists of a building, control center, loads, renewable  energy  resources, ESU, power grid,  and smart meter. The control center learns to take the optimal set of actions through interaction in a dynamic building environment with the goal of maximizing a certain reward quantity.}
\label{arch}
\end{figure*}
\subsection{Reinforcement Learning for EMS}
RL has gained increasing attention in EMS over the past decade \cite{raza2024smart}. Early methods relied on tabular Q-learning and discrete state representations \cite{7042790}, which limited scalability. Recent advances in deep RL enable EMS control in high-dimensional state and action spaces \cite{Multi-agent-havc,10677393,ren2022novel,huang2022mixed}. For example, \cite{ren2022novel} employed a forecasting-based dueling deep Q-network with a temperature-predicting LSTM module, while \cite{huang2022mixed} handled hybrid discrete-continuous actions. \cite{Multi-agent-havc,10677393} applied RL to control distributed buildings, but their approaches rely on idealized assumptions of perfect communication and coordination among buildings.

However, training deep RL agents directly in real buildings is often impractical due to high costs and safety concerns during exploration \cite{forootani2022advanced,zhang2023adaptivesafe}. To address this, model-based deep RL approaches have been introduced. For instance, \cite{9914662,10557792} leveraged EnergyPlus-generated data to build simulators for agent training, while \cite{arroyo2022reinforced} integrated MPC objectives into the RL framework using domain-encoded nonlinear controllers.
Despite these efforts, most existing approaches depend on accurate simulators or large-scale data collection, both of which are difficult in practice. Moreover, training separate agents for each building is costly and inefficient. Meta-learning offers a promising solution by enabling agents trained on a small number of buildings to generalize effectively to new environments.

\subsection{Meta-Reinforcement Learning}
Meta-RL aims to solve a new RL task by leveraging the experience learned from a set of similar tasks \cite{liu2021decoupling}. There are mainly two types of meta-RL algorithms: 
The first is recurrent-based meta-RL. 
In this case, the parameters of the prediction model are controlled by a learnable recurrent meta-optimizer and its corresponding hidden state \cite{10413635}. 
These approaches can achieve relatively good performance, but they may lack computational efficiency. 
The second is gradient-based meta-RL. These methods learn a well-generalized initialization that can be quickly adapted to a new scenario with a few gradient steps 
\cite{tancik2021learned,finn2017model}. 
Representatively, model-agnostic meta-learning (MAML) \cite{finn2017model} optimizes the initial policy network parameters of the base learner in the meta-training process, which significantly improves the efficiency of RL on the new task. 

\section{Preliminaries}


\subsection{Key Components in EMS}
A typical EMS architecture has several important components: building, control center, smart meter, loads, ESU,  renewable energy resources and power grid, 
as illustrated in Fig. 1. 
ESU could be lead-acid batteries or lithium-ion batteries, or a storage tank, which can reduce net-energy demand from main grids by storing excess renewable energies locally. Renewable energy resources could be solar panels or wind generators. Loads in an EMS can be generally divided into several types, e.g., non-shiftable loads, shiftable loads, non-interruptible loads, and controllable loads. To be specific, power demands of non-shiftable loads (e.g., televisions, microwaves) must be satisfied completely without delay. As for shiftable and non-interruptible loads (e.g., washing machines), their tasks can be scheduled to a proper time but can not be interrupted. In contrast, controllable loads (e.g., HVAC systems, heat pumps, and electric water heaters) can be controlled to flexibly adjust their operation times and energy usage quantities by following some operational requirements, e.g., temperature ranges. In this paper, we mainly focus on non-shiftable loads and thermostatically controlled loads. As for thermostatically controlled loads, HVAC systems are considered since they consume about 40\% of the total energy in a smart home. In each time slot, the 
control center makes the decision on ESU charging/discharging power and HVAC input power according to a set of available information (e.g., renewable generation output, non-shiftable 
load, outdoor temperature, and electricity price), with the aim of minimizing the energy cost of the EMS in the absence of the building thermal dynamics model.

\subsection{Markov Decision Process (MDP) Formulation}
In the building EMS, the indoor temperature at the next time slot is only determined by the indoor temperature, HVAC power input, and environment disturbances (e.g., outdoor temperature and solar irradiance intensity) in the current time slot. Moreover, the ESU energy level at the next time slot just depends on the current energy level and current discharging/charging power, which is independent of previous states and actions. Thus, both ESU scheduling and HVAC control can be regarded as a MDP. Assuming we are given a set of $N_t$ buildings
$B_S = {B_1, . . . , B_{N_t}
}$, then the sequential decision making problem associated in these buildings can be formulated as MDP as $M_{i}=<S_i,A_i,R_i, \gamma_i, P_i>$.  $S_i$ is the set of states $s_i$ and $A_i$ is the set of actions. $P_i$,  $R_i$ are the sets of state transition probabilities $p_i$ and rewards $r_i$; $\gamma_i\in[0,1]$ is a discount factor accounting for future rewards. In this paper, the agent denotes the learner and decision-maker (i.e., EMS agent), while the environment comprises many objects outside the agent (e.g., renewable generators, non-shiftable loads, ESU, the HVAC system, power grid, indoor/outdoor temperature). The EMS agent observes environment state $s^t$ and takes action $a^t$. Then, environment state becomes $s^{t+1}$ and the reward $r^{t+1}$ is returned, as shown in Fig. 1. 

Environment states in EMS MDP consist of seven kinds of information, i.e., renewable generation output $p^t$, non-shiftable power demand $b^t$,  outdoor temperature $T_{out}^t$ , electricity price $v^t$, ESU power demand $c^t$, HVAC power demand $h^t$  and time slot index  $t$.  Actions $A_i$ consist of the control on ESU and HVAC. 
We assume the power demand in EMS is always satisfied, which means the aggregated power supply should be equal to the served power demand. Then, we have $e^t  = b^t + h^t + c^t - p^t $ where $e^t$ is net electricity consumption.  

The agent aims to minimize the total energy cost and to do the load shaping to avoid the load of EMS fluctuating violently. Thus the corresponding reward consists of two parts, namely the penalty for the energy cost of the EMS, and the penalty for the changes in  electricity consumption in a short time. The first part of $R^t$ can be represented by $-C_{1}^{t}=-v^{t}*e^t$. Similarly, the second part of $R^t$ can be described by $-C_{2}^{t}=-\sum_{t'-W}^{t'}(e^{t}-e^{t-1})$, where $W$ is the window width of the electricity consumption we care about. The reward function $r_i^t$ can be represented as:
\begin{equation}
    r_i^{t} = -\mu C_{1}^{t} -\eta C_{2}^{t} ,
\end{equation}
where  $\mu$ and $\eta$ are  weighting coefficients of electricity cost and the changes of electricity consumption.

For each
building $B_i$, given an episode length $H_i$, the goal is to
learn an optimal control policy $\pi_i(a|s)$. In addition, the action-value function is defined as the sum
of reward $r_i^t$ discounted by $\gamma_i$ at each timestep $t$, which is formulated as 
\begin{equation}
 Q(s, a; f_{\theta}) = \mathbb{E} [r_i^t + {\gamma_i}r_i^{t + 1} + . . . |s_i^t = s, a_i^t = a].   
\end{equation}

Then, we defined the base learner $f$ with learnable parameter $\theta$ to map observations $S_i$ to outputs $A_i$. The effectiveness of function $f$ with optimal parameters $\theta_i$ is defined as
\begin{equation}
\begin{split}
   \mathcal{L}(f_{\theta_i})= & \\ \mathbb{E}_{s,a,r,s'\sim\mathcal{D}_i}
   & \left[(r + \gamma \max_{a'}Q(s',a'; f_{\theta^{-}_{i}})-Q(s,a;f_{\theta_i}))^2\right],     
\end{split}
\end{equation}
where $\theta^{-}_{i}$ are the parameters of target network that are fixed for every $C$ iterations.


\section{Proposed Solution}


\begin{figure*}[h]
\centering
 \setlength{\belowcaptionskip}{-0.3cm}
\includegraphics[width=.6\linewidth]{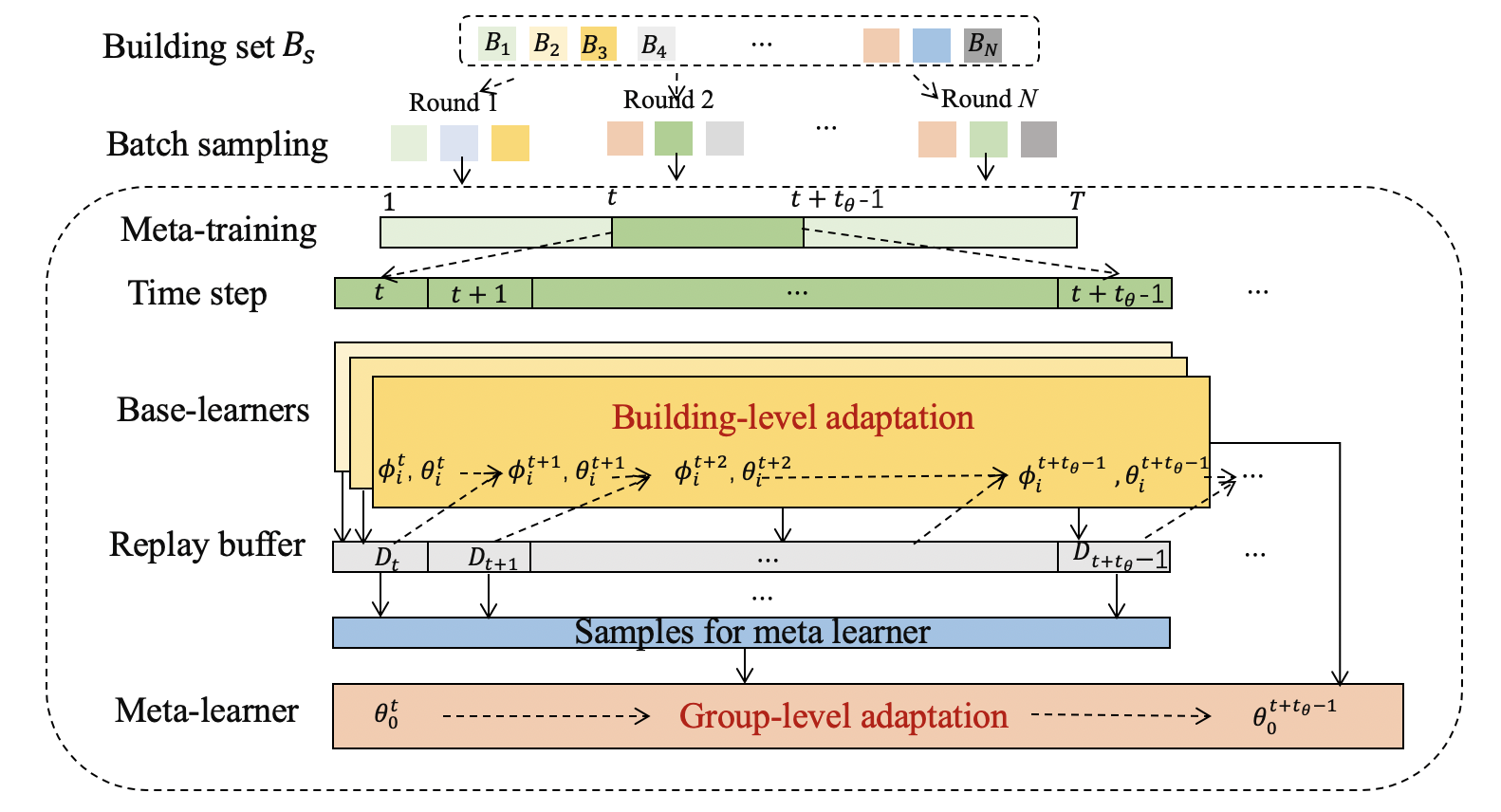}
\caption{Meta-training framework of \Name . From left to right, a batch of tasks are first sampled. Then, in meta-training, the whole episode with a length of T is split by $t_{\theta}$. During each interval $t_{\theta}$, the base-learner inherits the initialization from meta-learner and then conducts building-level adaptation using samples drawn from memory at each time step. At the end of each interval $t_{\theta}$, the meta-learner takes group-level adaptation with another batch of samples from the memory.}
\label{framework}
\end{figure*}
\subsection{\Name  Framework}
In \Name, we have a set of  $N_t$ training buildings, and each one is equipped with a base-learner $f$ to handle the scenario across different buildings. Besides, we have a well-generalized meta-learner $\mathcal{M}$ to enhance the learning efficiency of target unseen buildings.  As described in \cite{fujimoto2018addressing}, \Name  follows the design of the actor-critic method and the networks are parameterized by  $\theta$ (critic network) and $\phi$ (actor network). Besides, it is also equipped with experience replay and target network. During the training process, the parameters of base-learner $f$
(i.e., ${\theta_1, . . . , \theta_{N_t}
}$ and ${\phi_1, . . . , \phi_{N_t}
}$) and the well-generalized meta-learner
$\mathcal{M}(\cdot)$ are updated alternatively. The framework of \Name  is illustrated in Fig. \ref{framework}. 
\begin{algorithm}[t]
\caption{Meta-training process of \Name }
\label{alg:algorithm}

\begin{algorithmic}[1]
\Require{Set of source building $B_S$,
stepsizes $\alpha_\theta, \beta_\theta,\alpha_\phi, \beta_\phi$,
frequency of updating meta parameters $t_\theta$}
\Ensure{Optimized parameters initialization $\theta_0$, $\phi_0$}
\State Randomly initialize parameters $\theta_0$, $\phi_0$\
\For{round = $1,...,N$}
    \State Sample a batch of buildings from $B_S$\
    \For{$t =1 ,t_\theta+1, 2t_\theta+1,...,T$}
        \For{$t' = t,\min(t+t_\theta,T)$}
            \For{each building $B_i$}
                \State$\theta_i\gets \theta_0$, $\phi_i\gets \phi_0$
                \State Generate transitions into $\mathcal{D}$ and sample transitions as $\mathcal{D}_i$
                \State Update $\theta_i \leftarrow \theta_i-\alpha_\theta\nabla_{\theta}\mathcal{L}(f_{\theta_i};\mathcal{D}_i)$
                \State $\phi_i \leftarrow \phi_i-\alpha_\phi\nabla_{\phi}\mathcal{L}(f_{\phi_i};\mathcal{D}_i)$ by Eqn.(7) and Eqn.(8)
            \EndFor       
        \EndFor            
        \State Get and sample trajectories  $\mathcal{D}_i^{'}$ through each building's policy $\theta_i$
    \State Update  $\theta_0 \leftarrow \theta_{0}-\beta_{\theta}\nabla_{\theta}\sum_{B_i}\mathcal{L}(f_{\theta};\mathcal{D}_i^{'})$
    \State $\phi_0 \leftarrow\phi_0-\beta_\phi\nabla_{\phi}\sum_{B_i}\mathcal{L}(f_{\phi};\mathcal{D}_i^{'})$ by Eqn.(10) and Eqn.(11)
    \EndFor
\EndFor
\end{algorithmic}
\end{algorithm}
\subsubsection{Building-level Adaptation}
The building-level adaptation is a step-by-step optimization process of the individual buildings' task. 
For each building $B_i$, the agent’s experiences $
(s_i^{t}, a_i^{t}, r_i^{t}, s_i^{t+1})$ at each timestep $t$ are stored in set $\mathcal{D}_i$. The parameters
${\theta_1, . . . , \theta_{N_t}
}$ and ${\phi_1, . . . , \phi_{N_t}
}$ are learned by transitions $\mathcal{D}_i$ sampled
from each $B_i$. The goal is to minimize the loss over all buildings with base-learner $f_i$, which is defined as:
\vspace{-0.3em}
\begin{equation}
\{\theta_1, . . . , \theta_{N_t}\}^{'} := 
\argmin_{\{\theta_1, . . . , \theta_{N_t}\}}\sum_{i=1}^{N_t}\mathcal{L}(f_{\theta_i};\mathcal{D}_i).
\end{equation}
and
\vspace{-0.3em}
\begin{equation}
\{\phi_1, . . . , \phi_{N_t}\}^{'} := 
\argmin_{\{\phi_1, . . . , \phi_{N_t}\}}\sum_{i=1}^{N_t}\mathcal{L}(f_{\phi_{i}};\mathcal{D}_i).
\end{equation}
where the loss function of actor network $\mathcal{L}(f_{\phi_{i}})$  is:
\begin{equation}
    \mathcal{L}(f_{\phi_i})= -\sum_{s,a\in\mathcal{D}_i}Q(s,a;f_{\theta_i})
\end{equation}

In building-level adaptation, the
parameters $\theta_i, \phi_i$ of each task $T_{i_s}$ are updated at fixed timestep by gradient descent. They are formulated as (one gradient
step as an example):
\begin{equation}
    \theta_i \leftarrow \theta_i-\alpha_\theta\nabla_{\theta}\mathcal{L}(f_{\theta_{i}};\mathcal{D}_i),
\end{equation}
\begin{equation}
    \phi_i \leftarrow
    \phi_i-\alpha_\phi\nabla_{\phi}\mathcal{L}(f_{\phi_{i}};\mathcal{D}_i),
\end{equation}
where $\alpha_\theta,\alpha_\phi$ represents the step size updating the critic and actor network.

\subsubsection{Group-level Adaptation}
The group-level adaptation is a periodic synchronous updating process on a batch of sampled buildings' tasks. 
After the adaptation in
building-level, group-level adaptation aims to aggregate
the adaptation of each building $B_i$ using meta-learner $\mathcal{M}$, and then update
the initialization of  $\theta_0$ and $\phi_0$ in $\mathcal{M}$.
The meta-learner  $\mathcal{M}$  is optimized by several newly sampled batches of transitions $\mathcal{D}_i^{'}$ as follows:
\begin{equation}
    \mathcal{M}^{'}:= \argmin_\mathcal{M}\sum_{i=1}^{N_t}\mathcal{L}(\mathcal{M}(f_i);\mathcal{D}_i^{'})
\end{equation}

The meta-learner $\mathcal{M}$ is regarded
as well-generalized initialization of parameters in base-learner $f$. With a few gradient descent steps, we can
get the optimal parameters $\theta$. Thus, the meta-learner
$\mathcal{M}$ is regarded as
$\mathcal{M}(f_{\theta}) =  f_{\theta_0-\alpha\nabla_{\theta}\mathcal{L}(f_{\theta};\mathcal{D}_i^{'})}$. The whole loss in group-level training can be represented as $\mathcal{L}( f_{\theta_0-\alpha\nabla_{\theta}\mathcal{L}(f_{\theta};\mathcal{D}_i^{'})};\mathcal{D}_i^{'})$.

The initialization $\theta_0$ of $\mathcal{M}$ is updated as follows:
\begin{equation}
    \theta_0 \leftarrow \theta_0-\beta_\theta\nabla_{\theta}\sum_{B_i}\mathcal{L}(f_\theta;\mathcal{D}_i^{'}),
\end{equation}
in the same way, the initialization $\phi_0$ is updated as:
\begin{equation}
\phi_0 \leftarrow \phi_0-\beta_\phi\nabla_{\phi}\sum_{B_i}\mathcal{L}(f_{\phi};\mathcal{D}_i^{'}),
\end{equation}
where $\beta_\theta, \beta_\phi$ is defined as stepsize. 
The whole algorithm for the training process of \Name  is described in Alg. 1. Each task inherits a group-level initialization of parameters, then performs building-level adaptation and finally contributes to group-level adaptation. 


The \Name framework leverages prior knowledge to accelerate learning in unseen buildings, following the MAML-style gradient-based meta-RL paradigm. While traditional MAML focuses on policy-based DRL and only marginally outperforms random initialization in our setting, it falls short for real-world EMS deployment. To address this, \Name incorporates both building-level and group-level adaptation using an actor-critic method.
Additionally, unlike MAML, which updates policy parameters after full episodes, \Name performs adaptation at fixed timesteps to accelerate learning, allowing more frequent updates and reducing variance. Although control is applied per building, reward sharing enables inter-agent coordination, allowing buildings to learn cooperative behavior.


\subsection{Evaluation on New Buildings} 
In the meta-training process of \Name , we learn a well-generalized initialization of parameters in $\mathcal{M}$ with base-learners $f$. Then, we apply the initialization 
$\theta_0,\phi_0$ to a new target building $B_g$. By using $\theta_0,\phi_0$ as initialization, the update process in the building $B_g$ is defined as follows:
\begin{equation}
    \theta_g \leftarrow \theta_g-\alpha\nabla_{\theta}\mathcal{L}(f_{\theta};\mathcal{D}_g),
\end{equation}
\begin{equation}
    \phi_g \leftarrow \phi_g-\alpha_\phi\nabla_{\phi}\mathcal{L}(f_{\phi};\mathcal{D}_g).
\end{equation}
We evaluate the performance by using the optimal parameters $\theta_g,\phi_g$. The meta-testing process is outlined in Alg. 2.

\begin{algorithm}[t]
\caption{Meta-testing process of \Name }
\label{alg:algorithmtest}
\begin{algorithmic}[1]
\Require{Set of target buildings $B_T$; stepsizes $\alpha_\theta,\alpha_\phi$; learned initialization $\theta_0,\phi_0$}
\Ensure{Optimized parameters  $\theta_g,\phi_g$ for each building ${B_g}$ in ${B_T}$}
\State {Randomly initialize parameters $\theta_g$, $\phi_g$}
    \For{each building $B_g$ in $B_T$}
        \State{$\theta_g\gets \theta_0$}
        \State{$\phi_g\gets \phi_0$}
        \For{$t = 1,...,T$}
            \State Generate and sample  transitions as $\mathcal{D}_g$
            \State Update 
            $\theta_g \leftarrow \theta_g-\alpha_\theta\nabla_{\theta}\mathcal{L}(f_{\theta};\mathcal{D}_g)$
            \State $\phi_g \leftarrow \phi_g-\alpha_\phi\nabla_{\phi}\mathcal{L}(f_{\phi};\mathcal{D}_g)$ by Eqn.(12) and Eqn.(13)
        \EndFor
    \EndFor
\end{algorithmic}
\end{algorithm}

\section{Experiment}

\subsection{Experiment Settings}
We conduct experiments in an OpenAI-Gym style building energy management simulator called
CityLearn \cite{vazquez2019citylearn}, which provides the latest building EMS simulator to reshape the aggregated curve of electrical demand by controlling the energy applications of a diverse set of buildings. CityLearn includes energy models like the HVAC module: air-to-water heat pumps, electric heaters, space cooling, ESU module, the pre-computed energy loads of the buildings, and renewable resources like solar generation. We implement \Name  into CityLearn to control the HVAC and ESU. The RL agents send their control actions hourly and receive a set of states and rewards in return. The CityLearn environment came with one year of simulation data for building clusters from four climate zones. The energy demand for each building has been pre-simulated using EnergyPlus in a different climatic zone of the USA ( Hot-Humid : New Orleans;  Warm-Humid : Atlanta;  Mixed-Humid : Nashville;  Cold-Humid : Chicago) \cite{vazquez2019citylearn}. 
Assuming that the data distribution of buildings within the same climate zone is similar, we train and evaluate our control solution within the same climate zones and batch sample buildings from 8 building settings in each climate zone as training sets and use 3 randomly selected buildings as testing buildings, the training and testing set are disjoint. Then we repeat the experiment in four climate zones.




\textbf{Control Methods for Comparison}:
To evaluate the effectiveness of our \Name , we compare it with several representative methods described as follows.
\textbf{No Control}: a no control scenario, i.e., no-load shifting.
\textbf{RBC}: an RBC controller defined by CityLearn making decisions based only on the hour of the day. 
\textbf{Random initialization}: a soft actor-critic (SAC) \cite{haarnoja2018soft} agent defined by CityLearn using random initialization and training model from scratch.
\textbf{Pretrained}: a SAC agent randomly selects one existing model’s parameters as initialization for a new building. 
\textbf{MAML}: a SAC agent using MAML updating, which  mainly focuses on policy-based DRL problems, and it also conducts model updating at the end of a whole episode. 
\textbf{RL-MPC}: a method  \cite{arroyo2022reinforced} combines the MPC objective function with the RL agent value function while using a nonlinear controller model encoded from domain knowledge.

\textbf{Evaluation Metrics}: There are multi-metrics defined in the CityLearn to evaluate the performance EMS: 
\textbf{Net electricity consumption}  over the evaluation
period.
\textbf{1-load factor}: average monthly electricity demand
divided by its maximum peak.
\textbf{Ramping}: the district electricity
demand changes from one timestep to the next.
\textbf{Average daily peak} average daily peak demand over the evaluation period.
\textbf{Peak electricity demand}: over the evaluation
period.
For comparison convenience, all scores are normalized by the score of a benchmark RBC. A score of 0.96 means that
the RL controller performed 4\% better than the baseline RBC. 


In this work, SAC is chosen as base RL model for CityLearn environment. For our implementation of SAC, we
use a three-layer feedforward neural network of 64, 128 and 64
hidden nodes respectively, with rectified linear units (ReLU) between each layer for both the actor and critic and a final tanh unit following the output of the actor. 
Both network parameters are updated using Adam 
with a learning rate of $10^{-3}$. 
The batch size is 128 and $\gamma$ is 0.99. The target smoothing coefficient is 0.005 and the target update interval is 1. In \Name , the learning rates of building-learner and group-learner are set as $10^{-3}$ in both meta-training and meta-testing. The $\mu, \eta $ and $ W$ are set as 0.5, 0.5, and 5 respectively. 
In meta-training, the whole episode with a length of $T$ is split by $t_{\theta}=20$. During each interval $t_{\theta}$, the base-learner inherits the initialization from the meta-learner and then conducts building-level adaptation using samples drawn from memory at each time step. At the end of each interval $t_{\theta}$, the meta-learner takes group-level adaptation with another batch of samples from the memory. In RL-MPC, we follow the original paper  \cite{arroyo2022reinforced} and use a three-layer feedforward neural network to learn the control model and do the one-step prediction. We implement MAML with SAC following the traditional design, which mainly focuses on policy-based DRL problems, and it also conducts model updating at the end of a whole episode.
The training episode used in the Citylearn simulator is defined as a whole year for 8760 training timestep. The parameters $\theta$ and $\phi$ are initialized by sampling from a uniform distribution. 

\subsection{Experiment Results}

We train and evaluate our control solution on randomly new selected buildings sets, and then repeat the experiment in building
clusters in four climate zones. The control is performed for each building individually. We train the meta-learner of each climate zone separately. 

Tab. \ref{table-summary} and \ref{table-split}  show that \Name  performs best on all metrics given the limited training episode.
We report the average cost of our approach in comparison to other baselines in Tab. \ref{table-summary}, the cost is defined as the total non-negative net electricity consumption of the whole neighbourhood \cite{vazquez2019citylearn}. Considering the multi-metric have different timescale and measurements, each algorithm is evaluated on the testing set after one entire simulation episode. 
For control strategies with stochasticity, we report the mean and standard deviation of the cost over 5 random seeds. Most likely, since the cost is evaluated over the entire episode, the variance is small. 
From Tab. \ref{table-summary}, our approach consistently outperforms all our baselines. On average, we achieved a 5.45\% reduction in
average cost, compared to the benchmark RBC. The pattern of
the costs are consistent among four climate zones, indicating that our approach is robust to different climates. The breakdown of the overall cost of our approach by individual objectives after the entire episode is shown in Tab. \ref{table-split}. Note that, the improvement
is calculated by comparing with the best baseline.  Our approach performs particularly well in reducing ramping, average daily peak demand, annual peak demand and net electricity consumption, by 6.01\%, 13.58\%, 3.80\% and 2.02\% respectively. The 1-load factor is also better than others to remain a balanced load in the power grid.  

\begin{figure}[h!]
\centering
 \setlength{\abovecaptionskip}{-0.3cm}
  \setlength{\belowcaptionskip}{-0.2cm}
\includegraphics[width=.8\linewidth]{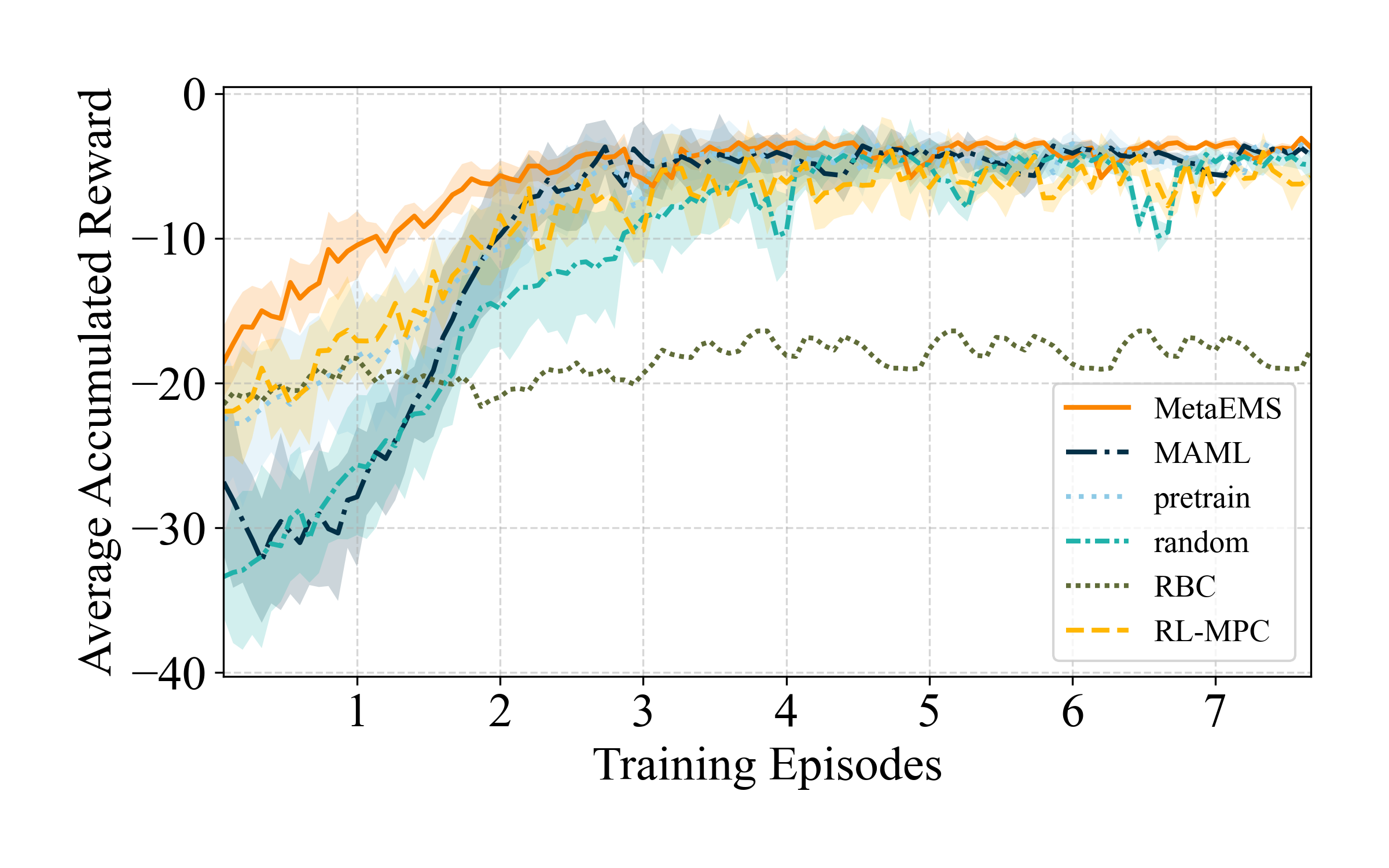}
\caption{ Accumulated average reward of buildings in climate zone 1 with means and variances of 5 random seeds. 
}
\label{figure-results-1}
\end{figure}

\begin{figure}[h!]
\centering
 \setlength{\abovecaptionskip}{-0.3cm}
 \setlength{\belowcaptionskip}{-0.2cm}
\includegraphics[width=.8\linewidth]{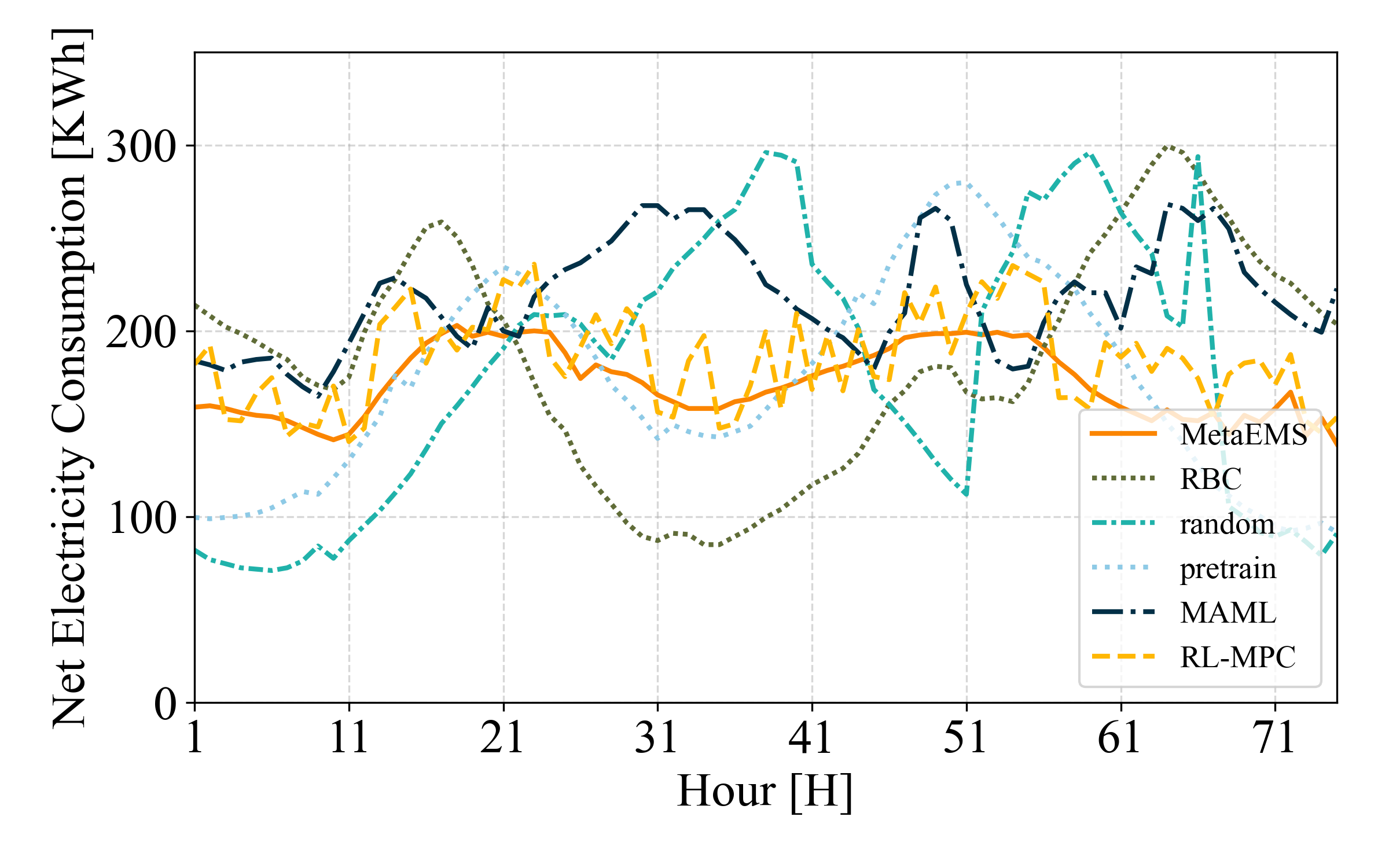}
\caption{Net electricity consumption over three days of a randomly selected building in climate zone 1.}
\label{figure-results-2}
\end{figure}

Fig. \ref{figure-results-1} and Fig. \ref{figure-results-2} show that \Name  maintains dramatically quicker and more stable adaptation in new buildings on electricity consumption than other baselines.   In Fig. \ref{figure-results-1}, we reused the building simulation data for more episodes in meta-testing, the random initialization, pretrained, MAML and RL-MPC  methods tend to converge around 7, 5, 5 and 6 episodes, respectively, and \Name  adapts quickly to the new building sets and converges in the first 3 episodes.
Besides, we can see that MAML and random initialization adapt very slowly and cannot keep a stable learning trend, which means the initialization that MAML has learned is no better than random initialization. Because of the high variance rooted in policy-based RL, the original policy-based updating mechanism MAML bring excessive instability to model updates. Thus, it is hard to learn a universal initialization in building control. In contrast,
Fig. \ref{figure-results-2} illustrates that the proposed methods can avoid violent fluctuation in the daily net electricity consumption curve compared with other baselines, since \Name penalizes the changes in electricity consumption in a short time. By incorporating task-based experience into model initialization, our approach is more sample efficient compared to the RL baselines training from scratch, MAML and pretrained, and can achieve lower electricity consumption than RL-MPC. 
With a real-world application as the end goal, it is essential that an algorithm does well with limited samples and timescale. Under the same episode, \Name  maintains a more stable and faster adaptation in new buildings. Incorporating building-level and group-level adaptation,
\Name  lets the base model learn more stably and efficiently by finding an optimal universal initialization.  

\begin{table}[t]
\renewcommand\arraystretch{1.3}
\centering
\caption{Summary of average cost. Each result is the average of three buildings.  }
\label{table-summary}
\resizebox{85mm}{!}{%
\begin{tabular}{lllll}
\hline
 &
  \multicolumn{1}{c}{\begin{tabular}[c]{@{}c@{}}Climate\\ 1\end{tabular}} &
  \multicolumn{1}{c}{\begin{tabular}[c]{@{}c@{}}Climate\\ 2\end{tabular}} &
  \multicolumn{1}{c}{\begin{tabular}[c]{@{}c@{}}Climate\\ 3\end{tabular}} &
  \multicolumn{1}{c}{\begin{tabular}[c]{@{}c@{}}Climate\\ 4\end{tabular}} \\ \cline{2-5} 
                 & \multicolumn{1}{c}{(\%)} & \multicolumn{1}{c}{(\%)} & \multicolumn{1}{c}{(\%)} & \multicolumn{1}{c}{(\%)} \\ \hline
            
No Control     & 109.34          & 114.13          & 100.73         & 105.19         \\       
RBC              & 100.00                   & 100.00                   & 100.00                   & 100.00                   \\ \hline
\makecell[l]{Random \\ initialization} 
& 101.18  $\pm$2.31                 & 103.59 $\pm$1.98            & 111.51   $\pm$2.96                & 104.26     $\pm$1.29               \\
MAML           & 103.91$\pm$0.86           & 98.91$\pm$0.45           & 99.12$\pm$2.42          & 97.02$\pm$1.50          \\
RL-MPC            &    98.22$\pm$0.59      & 100.53$\pm$0.41           & 98.67$\pm$1.69          & 101.20$\pm$3.51          \\
Pretrained       & 99.45$\pm$1.32           & 98.72$\pm$0.90           & 101.35$\pm$1.87          & 99.62$\pm$2.20           \\ \hline
\textbf{\Name } & \textbf{94.73$\pm$1.24}  & \textbf{92.65$\pm$0.81}  & \textbf{94.13$\pm$1.67}  & \textbf{96.65$\pm$0.82}  \\ \hline
\end{tabular}%
}

\end{table}


\begin{table}
\renewcommand\arraystretch{1.3}
\centering
\caption{Break down of cost by individual objectives on CityLearn environment. Each result is the average of three buildings and four climate zones. }
\label{table-split}
\resizebox{85mm}{!}{
\begin{tabular}{lccccc}
\hline
Metrics     &  \begin{tabular}[c]{@{}l@{}}Ramping\end{tabular} &   
\begin{tabular}[c]{@{}l@{}}1-load \\ factor\end{tabular}       & \begin{tabular}[c]{@{}l@{}}Avg. \\daily \\peak\end{tabular} & \begin{tabular}[c]{@{}l@{}}Peak\\ Elec. \\demand\end{tabular} & \begin{tabular}[c]{@{}l@{}}Net \\Elec. \\consumption\end{tabular}   \\  \hline
MAML              & 0.83          & 0.98          & 0.97              & 0.96             & 1.01                       \\
\makecell[l]{Random \\ initialization}           & 1.01          & 0.95          & 1.03              & 0.95             & 1.04                       \\
RL-MPC            & 1.00          & 0.99          & 1.02              & 0.97             & 1.02                       \\
Pretrained       & 0.99          & 0.94          & 0.98              & 0.97             & 
0.99\\
\textbf{\Name } & \textbf{0.78} & \textbf{0.98} & \textbf{0.83}     & \textbf{0.91}    & \textbf{0.98}              \\  \hline
Improvement      & 6.01\%        & $\setminus$       & 13.58\%            & 3.80\%           & 2.02\%                  \\
\hline
\end{tabular}}

\end{table}


\section{Conclusion}
In this paper, we proposed \Name, a meta-RL framework for efficient and adaptive energy management in building EMS.  \Name integrates both building-level and group-level adaptation mechanisms to learn a universal policy initialization that can rapidly adapt to new EMS scenarios without prior knowledge of system dynamics. Experiments in the CityLearn environment using real-world datasets demonstrate that \Name achieves faster convergence, shorter training time, and fewer environment interactions when adapting to unseen buildings. Moreover, the efficiency of \Name makes it a practical solution for real-world EMS applications where data and modeling resources are limited, such as battery storage management, renewable energy integration, and demand response in smart buildings.  In future work, we aim to explore the scalability of the framework in large-scale smart grid scenarios.


\bibliographystyle{IEEEtran}
\bibliography{IEEEabrv,ref}
\end{document}